# ADAPTIVE CHANNEL ENCODING FOR POINT CLOUD ANALYSIS

*Guoquan Xu[1], Hezhi Cao[2], Yifan Zhang[1], Jianwei Wan[1], Ke Xu[1], Yanxin Ma[1]*

[1]National University of Defense Technology, Changsha, CHINA
[2]University of Science and Technology of China, Hefei, CHINA
{xuguoquan19, zhangyifan16c, xuke, mayanxin}@nudt.edu.cn,
caohezhi21@mail.ustc.edu.cn, kermitwjw@139.com

**ABSTRACT**

Attention mechanism plays a more and more important role in point cloud analysis and channel attention is one of the hotspots. With so much channel information, it is difficult for neural networks to screen useful channel information. Thus, an adaptive channel encoding mechanism is proposed to capture channel relationships in this paper. It improves the quality of the representation generated by the network by explicitly encoding the interdependence between the channels of its features. Specifically, a channel-wise convolution (Channel-Conv) is proposed to adaptively learn the relationship between coordinates and features, so as to encode the channel. Different from the popular attention weight schemes, the Channel-Conv proposed in this paper realizes adaptability in convolution operation, rather than simply assigning different weights for channels. Extensive experiments on existing benchmarks verify our method achieves the state of the arts.

**Index Terms**—attention mechanism, channel attention, neural networks, adaptive channel encoding mechanism, channel-wise convolution

## 1. INTRODUCTION

Recently, point cloud processing technology is an urgent need in some applications such as robotics, autonomous driving and augmented reality [1]. Different from 2D images, point clouds have the characteristics of disorder and unstructured, which makes it very challenging to design neural networks to deal with them.

The initial work is to voxelize the point cloud or project it to 2D space [2-5]. These methods usually cause information loss, excessive consumption of memory, and high computation cost. PointNet [6] ensures invariance under displacement and rotation by using multi-layer perceptron (MLP), max-pooling and rigid transformation, creating a precedent for direct point cloud feature learning. Then PointNet++ [7] explores the local information aggregation method to improve the network performance. Affected by them, many recent works [8-11] consider defining convolution operators to aggregate the local features of point clouds.

Previous work focused on local feature extraction, ignoring the importance of different neighborhoods and channels. Therefore, attention mechanism has been proposed to solve this problem and achieved great success [12-16]. The attention network determines the contribution of each neighbor through the attention layer.

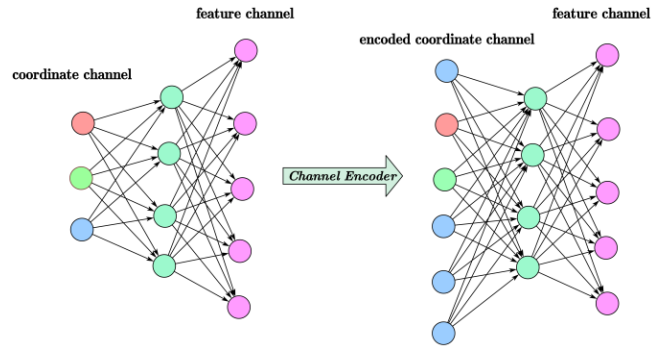

**Fig. 1.** Coordinate channel encoding. The contribution of coordinate channels to feature channels is learned by MLP (left). After channel encoding, the contribution difference of coordinate channels to feature channels is further mined (right).

Similarly, channel attention estimates a score function to weight the contribution of each channel.

A simple and effective channel encoding mechanism called Channel Encoder is proposed in this paper. It obtains the contribution of coordinate channels to each feature channel by Channel-Conv which can adaptively batch learn the relationship between them. Then, the representation ability of the network is improved by encoding the channel.

Specifically, all channels of coordinates contribute to feature extraction in general. However, in some cases, not each channel of coordinates is required. For example, if a 3D object is projected from a certain angle to 2D, people can still accurately classify it. That is, even if a channel is removed, it does not affect the judgment of object category. Hence, as shown in Fig. 1, Channel Encoder aims to learn the importance of each channel and retain the most important channel through channel coding. After channel encoding, the standard graph convolution is used for feature extraction.

The main contributions of this paper are summarized as follows:
- A channel encoding mechanism called Channel Encoder is proposed, which can encode each point to obtain the channel distribution more suitable for the related task.
- Channel-Conv is proposed, which can learn the relationship between coordinates and features, so as to gain the contribution of each channel. Then, the channel with the largest contribution is reserved and other channels are removed.
- Extensive experiments are exhibited on challenging benchmarks across three tasks including real-world

benchmarks, demonstrating our method achieves the state of the arts.

## 2. RELATED WORK

### 2.1. Point-based Deep Learning

PointNet [6] takes the lead in putting forward the idea of directly processing point cloud. And then, PointNet++ [7] tries to make up for the shortcomings of PointNet by capturing local structures through ball query grouping and hierarchical PointNet. Most of the follow-up works have been inspired by them and aim to design convolution operations on point clouds. EdgeConv is proposed in DGCNN [8] which can learn point relation in a high-dimensional feature space for capturing similar local shapes. However, the relationship learned by EdgeConv is not stable. A relation learning convolution operator RS-Conv [9] is proposed for RS-CNN. It can learn a high-level relation expression from geometric priors in 3D space. Moreover, it can explicitly learn the geometric relationship of points, which can improve shape perception and robustness. PAConv [10] proposes a position adaptive convolution operation with dynamic kernel assembling, which designs the convolution kernel by dynamically assembling basic weight matrices stored in Weight Bank. The coefficients of the weight matrices can be learned adaptively from point positions.

### 2.2. Attention mechanism for vision

Attention mechanism is initially applied to other areas, but it is very suitable for point cloud processing. Point cloud is unstructured and disordered and attention mechanism can deal with these problems well. Many works have introduced attention into vision tasks. A residual attention method with stacked attention modules is proposed for image classification in [12]. SE block is proposed for spatial encoding affected by attention mechanism [13]. SAGAN [14] is designed to apply self-attention for image generation. Some work has also introduced the attention mechanism into the point cloud analysis tasks. PCT [15] proposes a permutation-invariant point cloud transformer, which is suitable for learning unstructured point clouds with the irregular domain. Point Transformer [16] constructs self-attention networks with self-attention layers. Different from these methods, our method is to completely recode the channel, rather than simply assign the channel weight. Important channels are retained and their importance is amplified, while unimportant channels are removed.

## 3. METHOD

The cores of our method are Channel-Conv. In Sec. 3.1, the operation of Channel-Conv will be described in detail. Then the network structure will be introduced in Sec. 3.2.

### 3.1. Channel-Conv

A point cloud is denoted as $X = \{x_i | i=1,2,...,N\} \in \mathbb{R}^{N \times 3}$, and it has corresponding features $F = \{f_i | i=1,2,...N\} \in \mathbb{R}^{N \times C}$. Here, $N$ is the number of points and $C$ is the number of channel. $x_i$ represents the 3D coordinates of the i-th point. Sample the point $x_i$ as the central point and it has $k$ neighbors $\mathcal{N}(x_i)$ which can be obtained by k-nearest neighbors (KNN) or other methods.

Most of works only take the coordinates as input and obtain features from it. Thus, the subsequent features are closely related to the coordinates. Channel-Conv is designed to capture the potential relationship between their channels in this paper. As shown in Fig. 2, an adaptive channel convolution kernel is generated by features to encode the coordinate channel:

$$f_{ijk} = \Phi_k(f_i, f_j - f_i), \ x_j \in \mathcal{N}(x_i), \quad (1)$$

$$kernel = combine\{f_{ij1}, f_{ij2}, ..., f_{ijk}\}, \quad (2)$$

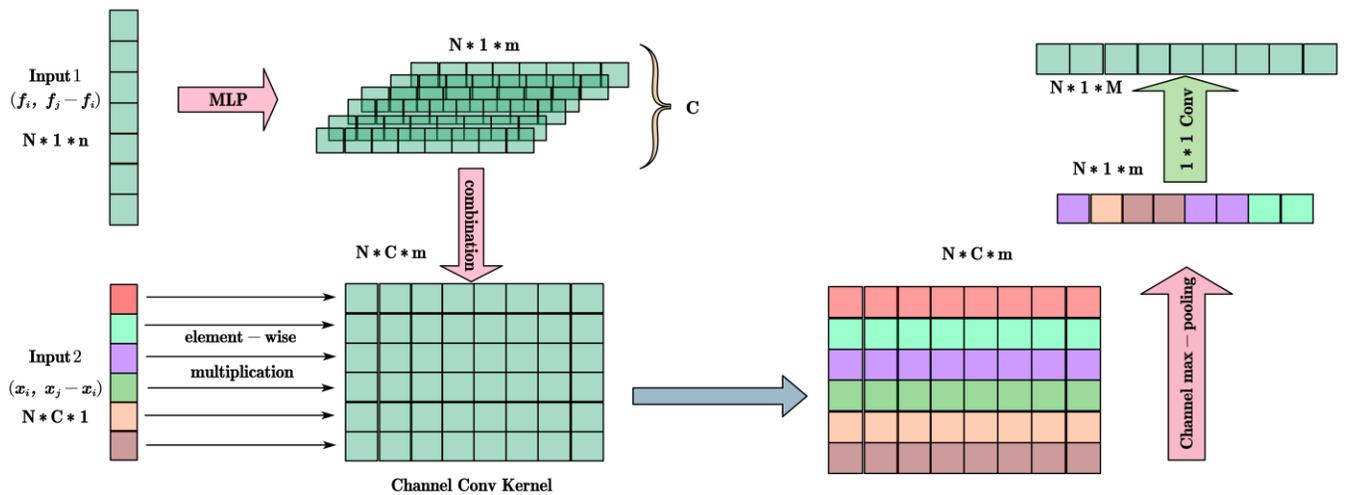

**Fig. 2.** The flow chart of Channel-Conv. Features are used to learn a channel convolution kernel, which convolutes each channel of coordinate. This is to learn the contribution of each channel to the feature, so element-wise multiplication is used instead of the dot-product. Then, the channel with the greatest contribution is retained through the max-pooling in the channel direction. Finally, a 1*1 convolution is implemented to extract features from the encoded channel as the feature input of the next layer.

where $\Phi(.)$ is a feature mapping function and implemented by MLP. $k = 1,2,\ldots,C$ denotes to learn the relevant convolution kernel for all $C$ coordinate channels. These convolution kernels are then combined to a channel convolution kernel. In order to better capture the local structure, $(f_i, f_j - f_i)$ is defined as the input feature for learning instead of $f_i$.

With the generation of channel convolution kernel, channel convolution is implemented:

$$\chi_{ij} = kernel \circledast \Delta x_{ij}. \tag{3}$$

Here, $\Delta x_{ij}$ indicates $(x_i, x_j - x_i)$ similar to feature input. $\circledast$ represents the element-wise multiplication of the corresponding rows of $kernel$ and $\Delta x_{ij}$. Because it is necessary to calculate the contribution of each channel of the coordinate relative to each channel of the feature. Using common dot product operation will produce channel aliasing, so it is impossible to measure their importance properly.

Since Eq. (3) obtains a contribution matrix, a max-pooling along the channel direction is adopted:

$$\Delta x'_{ij} = \max_{k \in C}(\chi^k_{ij}), \tag{4}$$

where $\max_{k \in C}()$ indicates max-pooling in the direction of the channel. It is the key to the final channel screening. The channels will be aliased if mean-pooling or sum-pooling is used.

Finally, a 1*1 convolution is used to extract features, and the neighborhood information is aggregated as the input of the next layer:

$$\Delta x'_i = \mathcal{A}\left(Conv(\Delta x'_{ij}), \forall x_j \in \mathcal{N}(x_i)\right), \tag{5}$$

where $\mathcal{A}$ is an aggregate function and $Conv$ means 1*1 convolution.

### 3.2. Network architecture

Two network architectures are designed to handle classification and segmentation tasks respectively. The network architectures are shown in Fig. 3. Our network can be roughly divided into three modules: Channel Encoder module, classification module and segmentation module.

**Channel Encoder.** It is the most important part which is composed of two Channel-Conv layers. The channel is reassigned to each point by two channel encoding layers. At the same time, the number of channels is increased, which is to better capture the importance of channels and obtain more accurate channel distribution.

**Classification network.** The classification network has two standard graph convolution layers and features of both of them are concatenated. There is no graph pooling between convolution layers. Since the coordinate channel has previously been encoded, the graph is updated by the similarity of features, rather than fixed spatial positions. This similarity is obtained by feature space distance:

$$d_{ij} = dis(f_i, f_j), \tag{6}$$

where $dis(.)$ means calculating the distance of the features and $d_{ij}$ is the feature space distance. To further capture high-level neighborhood relationships, the features are processed as follows:

$$f' = \Psi\left(\sigma(\psi(f))\right), \tag{7}$$

where both $\psi$ and $\Psi$ are mapping functions to learn higher semantic information. $\sigma$ indicates the activation function. Subsequently, the feature distance of higher dimensions can be calculated:

$$d'_{ij} = dis(f'_i, f'_j), \tag{8}$$

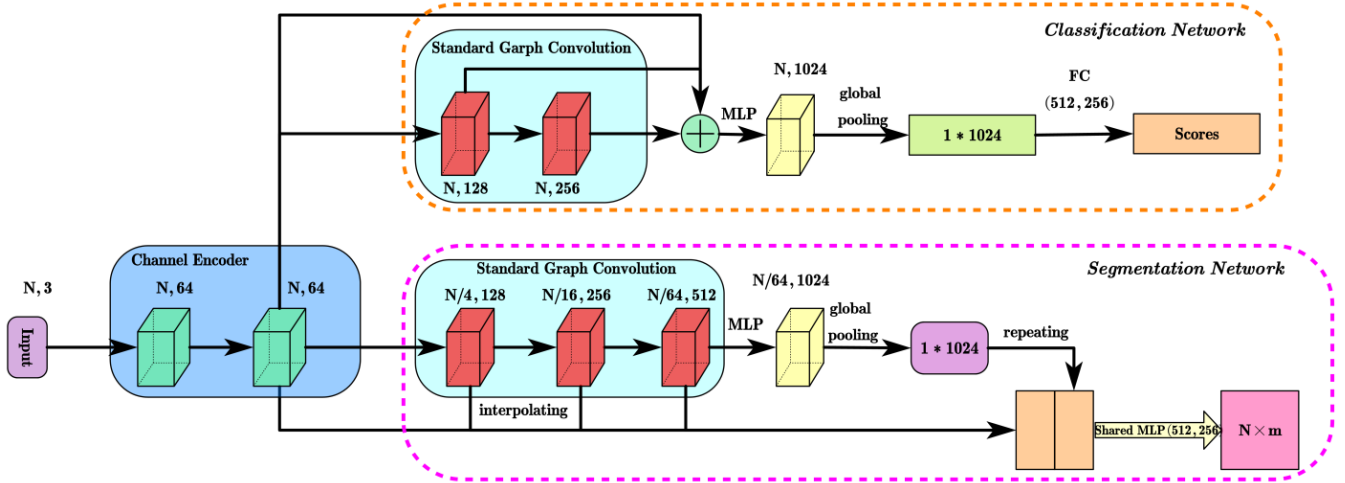

**Fig. 3.** Our network architectures for classification and segmentation tasks. Channel-Conv (green blocks) and standard graph convolution (red blocks) are applied to the architectures. Two Channel-Conv operations are used to encode the coordinate channel. The second layer is connected to the follow-up to extract important geometric information. After this, standard graph convolution is used for feature extraction. There are two standard graph convolution layers in the classification network and three in the segmentation network, and the features of all these layers are interpolated and concatenated for the final. The graph pooling is implemented only in the segmentation network.

These two feature distances are used in the classification tasks at the same time in this paper. The nearest $k$ points are used as adjacency points. The dynamic graph is constructed semantically, so that the receptive field of local neighborhood is expanded.

**Segmentation network.** There are three standard graph convolution layers in the segmentation network. The furthest point sampling algorithm is used to down sample the point cloud and then a new coarsened graph is constructed corresponding to the sampled points. The pooling layer is applied to get the aggregated features for points on the coarsened graph. Specifically, the max-pooling is implemented on the neighborhood of the sampled points, and gives the obtained features to the sampled points. In this way, a dynamic hierarchical structure can be built. In addition, interpolation and concatenation operations are applied for the final point features.

## 4. EXPERIMENTS

In this section, our method is evaluated on the idealized datasets ModelNet40 [17] and ShapeNet [18] and the non-idealized datasets ScanObjectNN [19]. The experimental details are provided. Afterwards, ablation studies and robustness experiments are implemented to verify the effectiveness of the modules and the performance of our method.

### 4.1. Shape Classification on ModelNet40

There are 9,843 train models and 2,468 test models in 40 classes in ModelNet40 [17] classification benchmark. The point cloud data is sampled from them by PointNet [6]. In this paper, 1024 points are sampled uniformly from each object and they are normalized to a unit sphere. Only coordinates are used as the input. The input data are augmented with random anisotropic scaling in the range [-0.66, 1.5] and translation in the range [-0.2, 0.2] during training.

The main parameter settings are as follows: the rate of dropout is set to 50% in the last two fully-connected (FC) layers; batch normalization and LeakyReLU are applied on all layers; the SGD optimizer with momentum set to 0.9 is adopted; The initial learning rate is donated to 0.1 and is dropped to 0.001 by cosine annealing. If there are no special instructions, these parameter settings are also adopted for other subsequent experiments.

**Table 1.** Classification accuracy (%) on ModelNet40. "vote" indicates voting tests and "no vote" indicates without voting tests. "w" means multi-scale inference and "no w" means without multi-scale inference.

| Method | Input | Accuracy |
|---|---|---|
| FPConv [20] | 1k points | 92.5 |
| PointCNN [11] | 1k points | 92.5 |
| KPConv [21] | 1k points | 92.9 |
| DGCNN [8] | 1k points | 92.9 |
| Grid-GCN [22] | 1k points | 93.1 |
| PCT [15] | 1k points | 93.2 |
| AdaptConv [23] | 1k points | 93.4 |
| RS-CNN (no w/ no vote) [9] | 1k points | 92.4 |
| RS-CNN (w/ no vote) [9] | 1k points | 92.9 |
| RS-CNN (w/ vote) [9] | 1k points | 93.6 |
| **Our method** | **1k points** | **93.6** |

The results of our method and the state-of-the-art methods on ModelNet40 are shown in Table 1. Our method achieves the best result when only 1k coordinates are used as the input. RS-CNN [9] achieves the same result as our method, but the result decreased to 92.9% without vote and 92.4% without multi-scale method. Our approach does not require these treatments.

### 4.2. Shape Part Segmentation on ShapeNet

The part segmentation experiment is implemented on ShapeNet [16]. The dataset is composed of 16,881 3D models from 16 categories. They are annotated with 50 parts in total and each point cloud is labelled with 2-6 parts. As in PointNet [6], each model is sampled 2048 points as the input.

**Table 2.** Shape part segmentation results (%) on ShapeNet.

| Method | Class mIoU | Instance mIoU |
|---|---|---|
| PointNet++ [7] | 81.9 | 85.1 |
| DGCNN [8] | 82.3 | 85.1 |
| SpiderCNN [24] | 81.7 | 85.3 |
| PointCNN [11] | 84.6 | 86.1 |
| PAConv (*DGC) [10] | 84.6 | 86.1 |
| RS-CNN [9] | 84.0 | 86.2 |
| AdaptConv [23] | 83.4 | 86.4 |
| **Our method** | **83.8** | **86.2** |

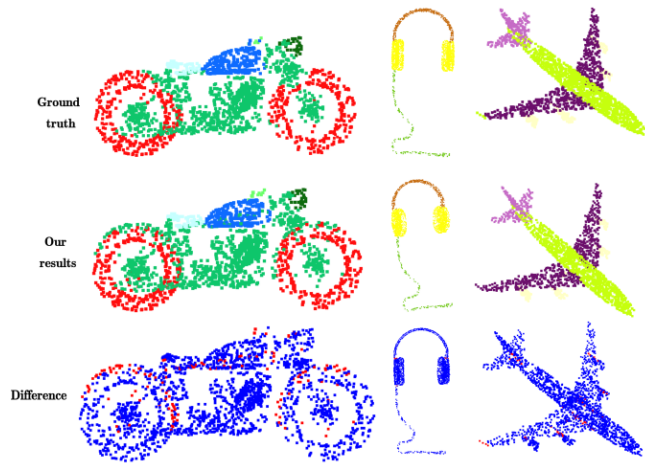

**Fig. 4.** Segmentation results on ShapeNet. The first row is the ground truth and the second row is our results. The error between our results and the ground truth is shown in the third row (red indicates the wrong point).

The quantitative comparisons with the state-of-the-art methods are reported in Table 2. Two types of mean Inter-over-Union (mIoU), class mIoU and instance mIoU, are used to compare the performance of these methods. Our method achieves satisfactory results. Visual segmentation results are shown in Fig. 4. The ground truth is shown in the first row and our results are shown in the second row. The third row shows the difference between them and blue represents correct and red represents wrong. It can be seen that the difference is very small.

### 4.3. Shape Classification on ScanObjectNN

ScanObjectNN [19] is adopted to further evaluate the performance of our method in this subsection. Different from ModelNet40 [17], ScanObjectNN is a real-world dataset obtained by scanning the real indoor scene. OBJ_ONLY, OBJ_BG, and HARDEST are its three subsets. OBJ_ONLY is the basic dataset which only includes the ground truth. OBJ_BG adds background interference to OBJ_ONLY. The HARDEST is obtained through a series of processing on the basis of OBJ_BG. These treatments include translating, rotating (around the gravity axis), and scaling the ground truth bounding box.

In order to be more convincing, our method is tested directly on the HARDEST and the results are summarized in Table 3. Our method is superior to other methods. AdaptConv [23] and RS-CNN [9] achieves 93.4% and 93.6% results on ModelNet40 [17], respectively. However, they are respectively reduced to 78.9% and 78.0% on the real dataset. At the same time, our method still performs well which makes our method have higher application value.

Table 3. Classification accuracy (%) on ScanObjectNN.

| Method | Hardest |
|---|---|
| PointNet [6] | 68.2 |
| SpiderCNN [24] | 73.7 |
| PointNet++ [7] | 77.9 |
| RS-CNN [9] | 78.0 |
| DGCNN [8] | 78.1 |
| PointCNN [11] | 78.5 |
| AdaptConv [23] | 78.9 |
| BGA-DGCNN [19] | 79.7 |
| BGA-PN++ [19] | 80.2 |
| **Our method** | **81.5** |

### 4.4. Ablation Studies

In order to verify the effectiveness of the proposed method, our Channel-Conv is replaced with Channel-wise Attention [25], Point-wise Attention [26] respectively. Besides, Channel-Conv is also replaced by standard graph convolution (GraphConv). For fairness, all parameters and network structure remain unchanged. The comparison results are reported in Table 4. From the table, our method is obviously superior to the other three schemes.

Table 4. The comparison results (%) on ShapeNet.

| Ablations | Class mIoU | Instance mIoU |
|---|---|---|
| GraphConv | 81.9 | 85.3 |
| Point-wise Attention | 78.1 | 83.3 |
| Channel-wise Attention | 77.9 | 83.0 |
| **Our method** | **83.8** | **86.2** |

As mentioned in Sec. 3.1, the pooling in the channel direction must be max-pooling, otherwise, channel aliasing will occur. The importance of this choice is explored through a set of experiments. Similarly, max-pooling is respectively replaced by mean-pooling and sum-pooling in the channel direction and other network architectures are identical. As shown in Table 5, the scheme of max-pooling achieves the best results.

Table 5. The comparison results (%) on ModelNet40.

| Ablations | Accuracy |
|---|---|
| mean-pooling | 93.4 |
| sum-pooling | 93.2 |
| **max-pooling** | **93.6** |

### 4.5. Robustness Experiments

The robustness of our method to point cloud density on ModelNet40 is further evaluated. Our method is trained with 1024 points and is respectively tested with sparser points of number 128, 256, 512, and 1024 as the input. The GraphConv and Channel-wise Attention mentioned in Sec. 4.4 are used as the comparison. From Fig. 5, compared with the other two methods, our method has higher robustness in the case of missing points.

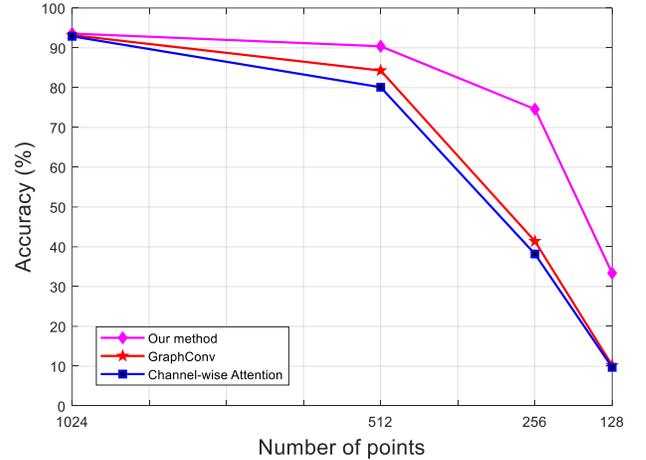

**Fig. 5.** The results of robustness experiments. It can be seen that our method is more robust. GraphConv indicates the standard graph convolution. Channel-wise Attention indicates the ablation where the Channel-Conv layers is replaced with graph attention layers (channel-wise).

### 5. CONCLUSION

In this paper, an adaptive channel encoding mechanism is designed for 3D point cloud analysis. The main contribution of our method is to design a Channel-Conv to adaptively learn the contribution of the channels and filter them. Instead of using a fixed convolution kernel, our method is dynamically generated by learning the relationship between features and coordinates. Our method has been implemented in multiple point cloud analysis tasks and achieved first-class results. Especially on the real-world dataset, our method outperforms the state-of-the-arts. In addition, Channel-Conv can be easily integrated into other graph CNNs to improve their performance.